# A Character-Level Neural Approach to Sinhala Sandhi Splitting

**Yasas Ekanayaka**
Informatics Institute of Technology,
Colombo 06, Sri Lanka
ydilshan.ek@gmail.com

**Deshan Sumanathilaka**
Informatics Institute of Technology,
Colombo 06, Sri Lanka
deshan.s@iit.ac.lk

## Abstract

Sinhala Sandhi splitting recovers the constituent words or morphemes hidden inside a phonologically merged surface form. The task is important for Sinhala NLP because Sandhi obscures lexical boundaries, but no prior published work has established a neural benchmark for Sinhala Sandhi splitting. We present a character-level sequence-to-sequence study based on SandhiLex, using native Sinhala Unicode input and evaluating recurrent encoder-decoder models for affixational and more complex lexicalized, derivational, and etymological Sandhi. The central challenge is the hard subset lexicalized, derivational, and etymological Sandhi, where our best model, a bidirectional LSTM encoder with a unidirectional LSTM decoder, reaches only 68.40% exact-match accuracy (82.08% character-level accuracy), well below the 94.00% achieved on the more regular affixational subset. Ablations show that bidirectional encoding is the largest contributor to performance, while native Sinhala script improves exact match accuracy over romanized input. Qualitative analysis indicates that many errors are near misses involving boundary adjacent characters or plausible but incorrect phonological substitutions. These results establish an empirical baseline for Sinhala Sandhi splitting and identify data scale, Sandhi type conditioning, and attention-based decoding as the main directions for future work.

## 1 Introduction

Most natural language processing pipelines assume that whitespace-delimited tokens correspond to meaningful lexical units. This assumption fails in morphologically complex languages where phonological processes at morpheme boundaries produce surface forms whose constituents are no longer recoverable by direct lexical lookup (Liyanage and Pushpananda, 2024). Sandhi, prevalent across South Asian languages, is a clear instance: adjacent words undergo vowel coalescence, consonant assimilation, or sound deletion upon combination, and the resulting errors propagate through downstream tasks, including morphological analysis, information retrieval, and machine translation (Ekanayaka et al., 2023).

Sinhala, an Indo-Aryan language spoken by over 16 million people in Sri Lanka, exhibits productive Sandhi across four linguistically distinct categories: affixational, lexicalized, derivational, and etymological (Liyanage and Pushpananda, 2024). For example, කලා kalā (art) and ආයතනය āyatanaya (institution) merge into කලායතනය kalāyatanaya (art institute), entirely obscuring the original word boundary. As Sinhala digital content continues to grow, the inability to correctly segment such forms increasingly limits the accuracy of downstream language technologies serving this community.

This gap is not only academic: everyday Sinhala software such as spell-checkers, dictionary lookup tools, and predictive typing systems depend on reliable morpheme boundary detection, and the absence of robust Sandhi splitting is a concrete obstacle to building such tools for Sinhala speakers today.

Existing computational approaches to Sinhala Sandhi rely on rule-based methods (Priyanga et al., 2017), which encode rules for generating merged forms rather than splitting them; reversing such rules for splitting offers no principled way to choose among multiple word pairs that could plausibly produce the same surface form, and these systems also require ongoing expert maintenance and generalize poorly beyond predefined rule inventories (V et al., 2014; Ghosh et al., 2022). Neural sequence-to-sequence models offer a data-driven alternative, achieving over 95% accuracy on Sanskrit Sandhi splitting (Aralikatte et al., 2018; Dave et al., 2020) with analogous successes for Tamil, Telugu, and Malayalam (Dasari et al., 2025; Pre-mjith et al., 2018). Despite the availability of SandhiLex (Liyanage and Pushpananda, 2024), a structured Sinhala Sandhi lexicon, and prior evidence that character-level neural models are effective for Sinhala morphological processing (Ekanayaka et al., 2023), no neural benchmark for Sinhala Sandhi splitting exists.

This gap raises three questions: which neural architectures are most effective for Sinhala Sandhi splitting; which factors, including encoder directionality, script representation, and data scale, most influence performance; and how quantitative metrics correlate with qualitative linguistic accuracy in this setting.

The present study addresses these questions by framing Sinhala Sandhi splitting as character-level sequence transduction (Dave et al., 2020; Dasari et al., 2025), training recurrent encoder-decoder models on both SandhiLex subsets. A bidirectional LSTM encoder with a unidirectional LSTM decoder proves most effective, achieving 68.40% exact match and 82.08% character-level accuracy on the hard subset and 94.00% on the affixational subset, with bidirectional encoding and native Sinhala Unicode identified as the largest contributors to performance. The contributions of this work are as follows:

1. The first neural benchmarks for Sinhala Sandhi splitting are established on both the affixational and hard subsets of SandhiLex.
2. Six recurrent encoder-decoder configurations alongside double-decoder and transformer variants are systematically compared under controlled conditions.
3. The first controlled comparison of native Sinhala Unicode versus romanized input is provided, showing that native script preserves orthographic distinctions critical for exact form recovery.

The remainder of this paper is structured as follows. Related works are reviewed, followed by the dataset and methodology. Results and analysis are then presented, and the paper concludes with a discussion of limitations and future directions.

## 2 Related Work

**Sinhala Sandhi and Sinhala NLP.** Sandhi splitting is a specific subtask within the broader field of morphological analysis: while morphological analysis more generally studies a word's internal structure, including inflectional and derivational paradigms, Sandhi splitting is concerned specifically with recovering constituent boundaries that have been obscured by phonologically conditioned surface changes at the point of merger. Sinhala Sandhi includes boundary changes such as vowel coalescence, consonant assimilation, insertion, and deletion (Liyanage and Pushpananda, 2024). SandhiLex distinguishes affixational, lexicalized, derivational, and etymological Sandhi.

Affixational Sandhi is more regular because it occurs at root affix boundaries; for instance, අසනීපයක asanipayaka (in an illness) splits cleanly into අසනීප asanipa (ill) + ය ya (ya) + ක ka (ka), where the boundary is transparent and components are recoverable by direct inspection. By contrast, etymological forms such as කම්මල kammala (smithy) offer no phonological cue to the correct split කම් kam (work) + හල hala (hall/place), as boundary consonants are assimilated and constituents no longer appear in their original form; rule based systems fail here by generating competing candidates without a principled selection mechanism. Sinhala morphological analysis has previously benefited from character-level neural models, including bidirectional GRU architectures, and identifies Sandhi splitting as an important next task (Ekanayaka et al., 2023). Similar character-level neural modelling has also been explored for Malayalam morphology and Sandhi-related suffix separation (Premjith et al., 2018; Sebastian and Kumar, 2020).

**Rule-based Sandhi processing.** Traditional Sandhi systems rely on manually specified phonological rules, finite-state analyses, or related rule-driven procedures (V et al., 2014; Ghosh et al., 2022). For Sinhala specifically, Priyanga et al. (2017) developed a rule-based word joiner that encodes phonological rules for combining word pairs into their merged Sandhi form. This system performs reliably for regular, rule-governed transformations such as vowel coalescence and consonant assimilation at predictable boundaries, and it does not require training data, making it practical for well-covered affixational patterns.

However, the system addresses *joining* (surface form generation) rather than *splitting* (surface form decomposition), and reversing a rule set built for generation is non-trivial when multiple underlying word pairs could plausibly produce the same merged form - a case that is common precisely in the lexicalized and etymological categories this study targets. Extending such a system to splitting would require an explicit disambiguation mechanism for competing rule applications, which the original system does not provide. Related rule-based studies in other South Asian languages show a similar pattern: linguistic rules are interpretable and require no training data, but they require expert maintenance and struggle when multiple underlying forms can produce similar surface strings (V et al., 2014; Kallur and Anami, 2025).

**Neural Sandhi splitting.** Most neural Sandhi work has focused on Sanskrit. Double Decoder RNNs, hybrid recurrent convolutional systems, and sequence-to-sequence models have reported strong results when trained on much larger Sanskrit resources (Aralikatte et al., 2018; Dave et al., 2020). Transformer and sequence-to-sequence approaches have also been explored for other Indian languages and technical lexicon generation settings (Dasari et al., 2025; J et al., 2025). These studies motivate a neural approach to Sinhala, but their data scale and linguistic settings differ substantially from those of SandhiLex: Sanskrit resources used in this prior work exceed 560,000 annotated examples, compared to the ~4,500 examples available in our hard subset. Because of this two-order-of-magnitude gap, we do not present a direct numerical comparison against these Sanskrit results in our evaluation, as doing so would risk attributing performance differences to architecture rather than to data availability. This study therefore evaluates simple, data efficient recurrent transducers before assuming that larger transformer style models are appropriate.

These observations motivate the experimental design of the present study. Character-level sequence-to-sequence modeling is adopted because it has proven effective for analogous morphophonological tasks in related low-resource languages (Premjith et al., 2018; Dave et al., 2020). Bidirectional encoding is prioritized given prior evidence for Sinhala morphological processing (Ekanayaka et al., 2023); Section 6 shows this benefit is capacity-dependent rather than automatic. The evaluation is conducted separately on affixational and hard subsets to isolate the effects of linguistic complexity and data scale, following practices established in Sanskrit Sandhi research (Aralikatte et al., 2018; Sreedeepa et al., 2024).

## 3 Task and Data

Given an input Sandhi form $x = x_1, \ldots, x_n$, the task is to generate a target sequence $y = y_1, \ldots, y_m$ containing the split constituents. For instance, විහාරාධිකාර viharādhikāra (temple authorities) is mapped to විහාර vihāra (temple) + අධිකාර adhikāra (authority). The output may be longer than the input because splitting can require restoring characters that were changed or deleted during Sandhi formation. Table 1 summarizes the Sandhi categories used in the dataset with examples.

The experiments use SandhiLex, a Sinhala Unicode Sandhi lexicon developed and released by Liyanage and Pushpananda (2024) at the Language Technology Research Laboratory (LTRL), University of Colombo School of Computing, to formalize Sinhala morphophonology for computational use. SandhiLex is not introduced in this work; it was obtained for this study through a formal academic request to LTRL-UCSC, and the present paper's contribution is the first neural modeling study built on top of it, not the resource itself. The dataset is distributed as plain text tab-separated files in which each row pairs one merged Sandhi form with its segmented constituent form. In this study, each row is treated as a supervised sequence pair: the Sandhi word is the encoder input, and the split form is the decoder target.

Following the experimental setup, we evaluate two subsets. The *easy* subset contains Affixational Sandhi, where the boundary occurs between a lexeme and an attached prefix or suffix. This subset contains more than 300,000 available pairs; 150,000 are sampled for the full data experiment (Table 2). Because Affixational Sandhi is comparatively systematic, it is used to test how the model behaves when many regular examples are available.

Table 2: SandhiLex subsets used in the experiments. The easy subset is affixational; the hard subset combines lexicalized, derivational, and etymological forms.

| Subset | Sandhi types | Size |
|---|---|---|
| Easy | Affixational | >300k pairs; 150k sampled |
| Hard | Lexicalized, derivational, etymological | ~4.5k pairs |

The *hard* subset contains approximately 4,500 Lexicalized, Derivational, and Etymological

Table 1: Sinhala Sandhi categories used in SandhiLex, with transliteration and English glosses.

| Type | Description | Example input | Split |
|---|---|---|---|
| Affixational | Root affix boundary; comparatively regular | අසනීපයක asanipayaka (in an illness) | අසනීප ය ක asanipa ya ka (ill + suffixes) |
| Lexicalized | Word combination whose meaning remains related to its components | දකුණත dakunata (right hand) | දකුණු අත dakunu ata (right + hand) |
| Derivational | Boundary change creates a derived lexical form | ලේඛනාගාර lekhanāgāra (archives) | ලේඛන ආගාර lekhana āgāra (writings + house) |
| Etymological | Historically fused form with less transparent components | කම්මල kammala (smithy) | කම් හල kam hala (work + hall/place) |

Sandhi forms collected through semi automatic manual and corpus based extraction (Liyanage and Pushpananda, 2024). These forms are harder because the original boundary is often less transparent and the surface form may not preserve all characters from the underlying constituents. The hard subset is therefore the main benchmark for evaluating whether the model can handle opaque Sinhala Sandhi. Examples include දකුණත dakunata (right hand), split as දකුණු dakunu (right) + අත ata (hand), and කම්මල kammala (smithy), split as කම් kam (work) + හල hala (hall/place).

**Data format** SandhiLex is distributed as plain-text files encoded in Sinhala Unicode, with each entry separated by a single tab character: `<sandhi word>\t<split form>`. For example, a row contains a single merged input token on the left and a whitespace-separated sequence of recovered components on the right. This tab-separated structure is preserved through preprocessing, where each row is treated directly as an encoder-input/decoder-target pair.

**Romanized counterpart.** For the script representation experiment, every native Unicode pair is converted into a romanized pair while preserving the same input target alignment. The transliteration module uses a character-level mapping in which Sinhala symbols are categorized as consonants, modifiers, or vowels. Consonants carry an implicit *a* vowel unless a modifier changes or suppresses it. For example, the native pair අක්ෂාංශ akshāmsha (latitude/part) → අක්ෂ අංශ aksha amsha (axis/letter + part) is represented in the romanized data with the same segmentation boundary. This controlled pairing allows differences in performance to be attributed to script representation rather than to a different train-test split (Ekanayaka et al., 2023).

# 4 Methodology

## 4.1 Preprocessing

Before training, the raw SandhiLex files are cleaned by removing empty and duplicate entries. The Sandhi word is used as the encoder input, and the segmented form is used as the decoder target. The target sequence is bounded by a start token ($) and an end token (¶). Character-level tokenisation is applied separately to encoder and decoder sequences, and sequences are post-padded to uniform lengths for batch training. Deep learning is applied to both the easy and hard subsets in this study, rather than reserving neural models for the hard subset alone. This is a deliberate choice: although affixational (easy) Sandhi is comparatively regular, rule-based methods still require ongoing manual maintenance and generalize poorly to forms outside their predefined rule inventories. Applying rule-based methods to the easy subset would reintroduce these limitations, so both subsets are evaluated within a single, consistent neural framework.

Native Sinhala Unicode is the primary representation. However, romanized transliteration is sometimes used in low resource NLP to reduce character vocabulary size and simplify tokenization, motivating a controlled comparison. A romanized version of the same data is prepared using a character-level mapping that categorizes Sinhala symbols as consonants, modifiers, or vowels, preserving identical Sandhi and split pairings so that any performance difference is attributable to script choice rather than dataset variation. The key concern is that romanzation may merge phonologically distinct Sinhala characters into identical Roman forms, losing the orthographic contrasts that mark Sandhi boundaries. This follows evidence that native Sinhala script preserves orthographic distinctions important for morphological process-ing (Ekanayaka et al., 2023).

### 4.2 Encoder-Decoder Model

Figure 1 illustrates the overall Architecture of the encoder-decoder system. A merged Sinhala Sandhi form is tokenised at the character level and fed into the encoder as a sequence of integer indices. The bidirectional LSTM encoder processes the sequence in both forward and backward directions with latent dimensionality 512 per direction; the resulting forward and backward hidden and cell states are concatenated to form a 1024 dimensional context vector that summarises the full input. This context vector initialises the unidirectional LSTM decoder, which autoregressively generates one output character per step through a dense softmax projection until the end token or maximum length is reached, producing the whitespace separated split form.

The model is compiled with the Adam optimizer and sparse categorical loss. During inference, separate encoder and decoder models are instantiated: the encoder produces the initial decoder states, and the decoder generates the segmented output autoregressively until the end token or maximum output length is reached.

### 4.3 Training Protocol

Separate models are trained for the easy and hard data settings. The easy dataset is trained for 10 epochs with batch size 32. The hard dataset is trained for 25 epochs owing to its smaller size and greater linguistic complexity. A 10% validation split is used during training, and final evaluation follows an 80:20 train-test setup. Tokenizers, vocabulary mappings, model parameters, and sequence length settings are serialized for reproducible inference. The model and training settings are summarized in Table 3.

Table 3: Implementation settings for the main model.

| Component | Setting |
|---|---|
| Input unit | Sinhala Unicode character |
| Output unit | Sinhala Unicode character plus space |
| Encoder | Bidirectional LSTM, 512 units per direction |
| Decoder | Unidirectional LSTM, 1024 units |
| Loss | Sparse categorical cross-entropy |
| Optimizer | Adam |
| Easy training | 10 epochs, batch size 32 |
| Hard training | 25 epochs, batch size 32 |
| Validation | 10% validation split |
| Final split | 80:20 train-test |
| Decoding | Autoregressive |

### 4.4 Compared Systems and Metrics

The study compares RNN, GRU, and LSTM encoder-decoder models with both unidirectional and bidirectional encoders, following recurrent modeling choices in Malayalam and Sanskrit work (Premjith et al., 2018; Hellwig and Nehrdich, 2018). Additional experiments include a manually configured LSTM encoder-decoder, Double Decoder LSTM variants inspired by Sanskrit work (Aralikatte et al., 2018), and a SinBERT Large transformer fine-tuning experiment.

Exact match (EM) measures the percentage of predictions that exactly match the reference split. Character-level accuracy (CLA) measures the agreement between predictions and references at the character level. Word-level precision, recall, and F1 are reported on a randomly selected 500-sample hard set test subset to provide a more fine-grained evaluation of token-level prediction quality, particularly for partially correct outputs and boundary-sensitive errors.

## 5 Results

Table 4 shows the main hard subset architecture comparison. Bidirectional encoders outperform unidirectional encoders for all recurrent cells. The best model is the bidirectional LSTM encoder-decoder, with 68.40% EM and 82.08% CLA. Relative to the unidirectional LSTM, bidirectionality adds 26.0 EM points, consistent with the capacity-dependent pattern analyzed in Section 6.

Table 4: Architecture comparison on the SandhiLex hard subset. Scores are percentages.

| Encoder | Cell | EM | CLA |
|---|---|---|---|
| Unidirectional | RNN | 20.40 | 56.44 |
| Unidirectional | GRU | 33.40 | 66.53 |
| Unidirectional | LSTM | 42.40 | 71.50 |
| Bidirectional | RNN | 24.40 | 62.01 |
| Bidirectional | GRU | 50.50 | 74.59 |
| Bidirectional | LSTM | **68.40** | **82.08** |

To assess whether the BiLSTM improvement over BiGRU is statistically reliable rather than an artefact of test set variation, McNemar's test is applied to the paired predictions on the randomly selected 500 sample hard set subset. McNemar's test is appropriate here because the two models are evaluated on the same examples, making the errors dependent rather than independent, which violates the assumptions of standard proportion tests.

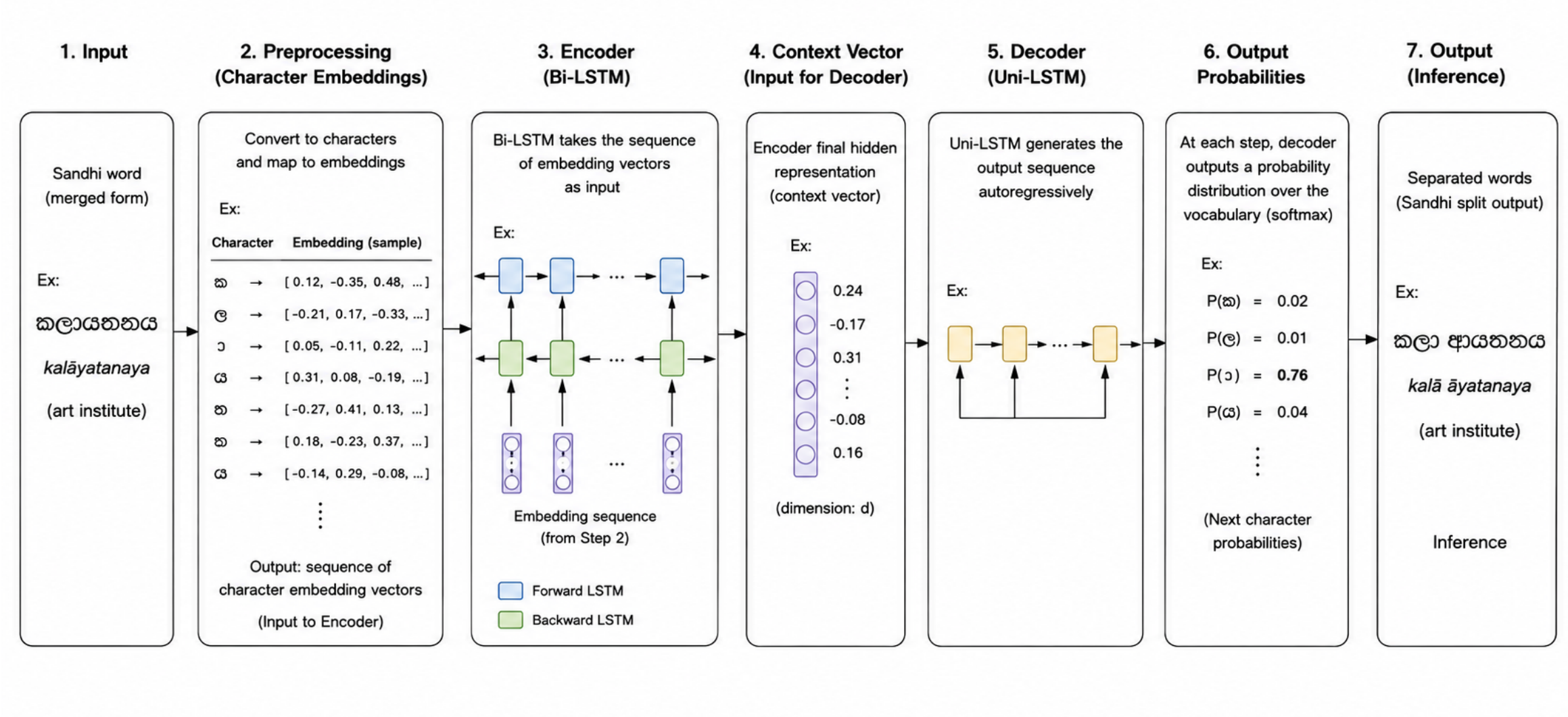

Figure 1: Architecture of the bidirectional LSTM encoder-decoder for Sinhala Sandhi splitting

The result is significant $3.076 \times 10^{-7}$ $(p < 0.05)$, confirming that the BiLSTM's advantage over BiGRU reflects a genuine difference in predictive behaviour and is not attributable to chance, as summarized in Table 5.

Table 5: Significance test reported for the two strongest recurrent encoders.

| Comparison | Result |
|---|---|
| BiLSTM vs. BiGRU | $3.076 \times 10^{-7}$ $(p < 0.05)$ |
| Test subset | 500 examples |
| Test type | McNemar |

**Gap between EM and CLA.** The consistent gap between EM and CLA across all models indicates that many incorrect predictions are near misses rather than complete failures. For instance, දකුණත dakunata (right hand) is predicted as දකුණ අත dakuna ata (right hand) instead of the correct දකුණු අත dakunu ata (right hand), recovering the correct constituents and boundary position but missing a single final vowel.

Similarly සිහසුන sihasuna (the throne), the model produced සිහු අසුන sihu asuna (lion seat) instead of the correct සිහ අසුන siha asuna (lion seat), again preserving the boundary and overall structure while omitting a final vowel. Such errors score zero under exact match yet achieve high character-level overlap, a pattern consistent with observations in Sanskrit Sandhi segmentation (Sreedeepa et al., 2024). This motivates reporting both metrics rather than relying on a single aggregate score.

Table 6 compares native Sinhala Unicode with romanized input for bidirectional encoders. The best exact-match score is obtained with native-script LSTM input. Romanization gives higher CLA for LSTM, but lower EM, suggesting that romanized forms may ease local character prediction while losing distinctions needed for complete form recovery.

Table 6: Script representation comparison for bidirectional encoders on the hard subset. Scores are percentages.

| Script | Cell | EM | CLA |
|---|---|---|---|
| Romanized | RNN | 31.80 | 73.78 |
| Romanized | GRU | 58.75 | 81.92 |
| Romanized | LSTM | 64.00 | **86.27** |
| Native Unicode | RNN | 24.40 | 62.01 |
| Native Unicode | GRU | 50.50 | 74.59 |
| Native Unicode | LSTM | **68.40** | 82.08 |

Table 7 separates data scale from linguistic difficulty. With the full affixational subset, the bidirectional LSTM reaches 94.00% EM and 98.24% CLA. When the easy subset is size matched to the hard training regime, EM falls to 68.40%. This supports the study conclusion that the hard subset is constrained not only by linguistic opacity, but also by limited data volume, a recurring concern in low resource and Sandhi specific research (de Silva, 2019; Sreedeepa et al., 2024).

Table 7: Effect of data volume and Sandhi type on bidirectional LSTM performance.

| Dataset | Condition | EM | CLA |
|---|---|---|---|
| Easy | Size matched | 68.40 | 78.85 |
| Easy | Full 150k sample | **94.00** | **98.24** |
| Hard | Full ∼4.5k | 68.40 | 82.08 |

Figure 2 summarizes the additional model families evaluated on the hard subset. The Double Decoder LSTM was implemented following the architecture proposed by (Aralikatte et al., 2018) for Sanskrit, and adapted to Sinhala by training separate encoder and decoder components on SandhiLex, using the same 80:20 train-test split and 25-epoch training protocol as for the main models.

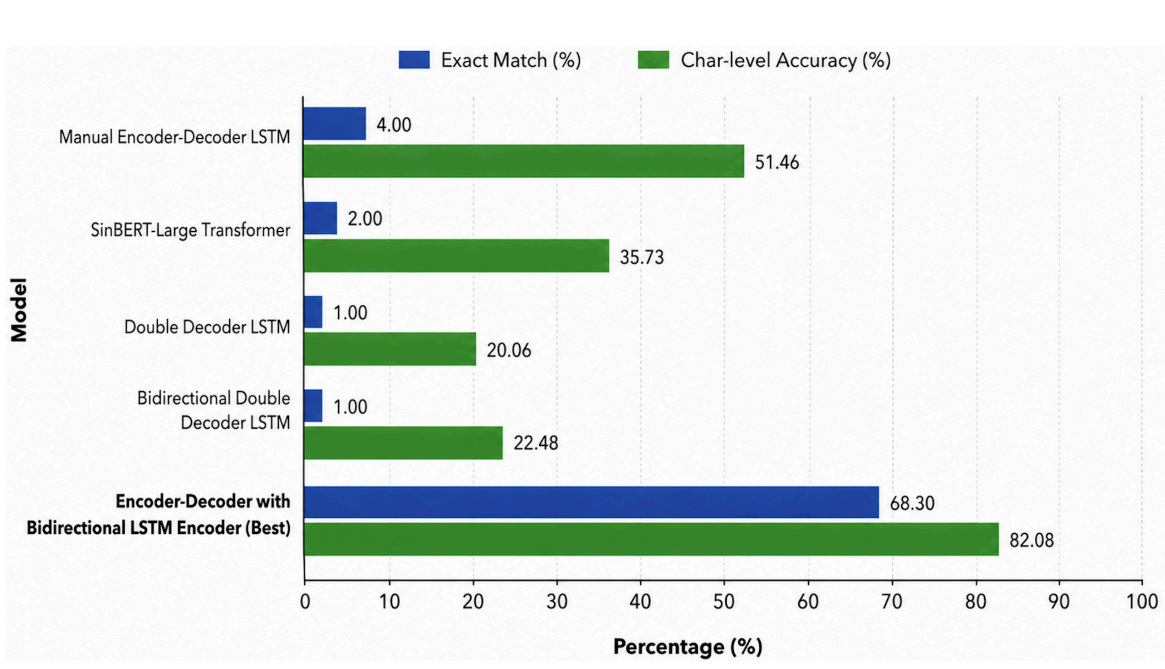

Figure 2: Additional architecture results on the hard subset. Scores are percentages.

The SinBERT Large experiment fine-tuned the pre-trained SinBERT Large transformer on the hard subset using the same character-level formulation. Neither SinBERT Large nor the Double Decoder LSTM outperformed the recurrent encoder-decoder baseline, achieving only 2.00% and 1.00% EM, respectively.

These results suggest limited transferability of the Sanskrit-oriented Double Decoder architecture to Sinhala under low-resource conditions, while the SinBERT Large result is consistent with findings that transformer models can be less data-efficient in small-data settings (Sreedeepa et al., 2024). However, these findings should be interpreted only within the current dataset and implementation setup (Aralikatte et al., 2018; Dasari et al., 2025).

On the randomly selected 500 sample hard subset test set, word level exact match precision, recall, and F1 are all 63.60%. This equality follows from the exact match evaluation setup, where predictions are counted as either fully correct or incorrect.

## 6 Analysis and Discussion

The results identify the bidirectional LSTM encoder with a unidirectional LSTM decoder as the strongest architecture for Sinhala Sandhi splitting, consistent with earlier evidence supporting bidirectional character-level modelling for Sinhala morphology (Ekanayaka et al., 2023). The model achieves 68.40% EM, 82.08% CLA, and 63.60% F1 on the hard subset, outperforming both the unidirectional LSTM and bidirectional GRU baselines. Native Sinhala Unicode input also improves exact recovery compared to the romanized LSTM setting.

The comparison between unidirectional and bidirectional encoders shows that two-way context is an important design factor, but the size of its benefit is not uniform across cell types: bidirectionality adds 26.0 EM points for LSTM, 17.1 points for GRU, and only 4.0 points for RNN (Table 4). This pattern suggests that access to right-context alone is not sufficient - the encoder also needs enough gating capacity to retain and combine information from both directions over the length of a Sandhi form, which plain RNNs lack due to their limited ability to preserve long-range state. This matches the linguistic nature of Sandhi, where boundary transformations depend on characters on both sides of the merge (Liyanage and Pushpananda, 2024), but indicates that the benefit is conditional on architectural capacity rather than automatic from bidirectionality alone. The script experiment also supports the conclusion that native Sinhala Unicode is preferable for exact form recovery, even though romanized input can sometimes achieve higher character-level accuracy.

The easy and hard comparison highlights the role of both data volume and Sandhi type. Affixational Sandhi reaches 94.00% EM in the full 150,000 sample setting because its boundary patterns are highly regular. In contrast, the hard subset reaches 68.40% EM because lexicalized and etymological forms involve more irregular phonological changes and are trained on only ∼4,500 examples. When the easy subset is size matched to the hard setting, its EM also drops substantially, confirming that data volume contributes significantly to the performance gap (Liyanage and Pushpananda, 2024).

The additional architecture experiments show that the tested SinBERT Large and Double Decoder LSTM settings do not outperform the bidi-rectional LSTM encoder-decoder under the available data conditions, despite the success of transformer and double-decoder approaches in other Sandhi settings (Aralikatte et al., 2018; Dasari et al., 2025). These findings are interpreted as results specific to the current dataset and implementation setup rather than as a general limitation of such architectures.

## 7 Ablation Studies

Encoder directionality, the single largest contributor to performance, is already isolated by the unidirectional-versus-bidirectional LSTM comparison in Table 4 (Section 5), where bidirectionality adds 26.0 EM points (42.40% to 68.40%); it is therefore not re-ablated here. Three additional factors - native script, latent dimension, and hyperparameter tuning - were ablated separately from the full model; full results are reported in Appendix A (Table 10).

Each of these three factors contributes meaningfully but less than encoder directionality: reduced latent dimension has the largest effect of the three, followed by removing hyperparameter tuning and then romanizing the input (see Appendix A for exact deltas). Although romanization improves CLA, native Sinhala Unicode remains preferable for exact segmentation because it preserves orthographic distinctions needed for complete form recovery.

## 8 Error Analysis

The error analysis identifies three recurring error types. Minor boundary shifts occur when the model finds the approximate split location but alters a nearby character (සිහසුන sihasuna (throne) → සිහු අසුන sihu asuna (lion seat) instead of සිහ අසුන siha asuna (lion seat)). Phonological substitutions produce plausible but wrong decompositions (සමීකරණ samīkarana (equation) → සං කරණ sam karana (co-doing/formation) instead of සම කරණ sama karana (equal making)). Complete segmentation failures occur mostly for low-frequency etymological compounds: for කම්මල kammala (smithy), the model predicts කම් මල kam mala (work flower), an orthographically plausible but semantically wrong split, instead of කම් හල kam hala (work hall/place) (Table 8).

These three error types are not equally fixable. Boundary shifts and phonological substitutions occur where the correct split remains phonologically plausible from the surface characters, so additional context or model capacity could plausibly reduce them. Complete failures on etymological compounds are qualitatively different: forms such as කම්මල kammala (smithy) retain no surface trace of the historically original හල hala (hall/place), so no amount of additional context can recover it from the input alone. Rule-based systems fare no better here, since they lack any principled way to disambiguate among competing surface-plausible splits either. This suggests etymological Sandhi is better modeled as a lexical retrieval problem – matching known fused forms to their historical constituents – rather than as purely context-driven sequence transduction, which is consistent with the capacity ablation (Appendix A) narrowing the gap for boundary-shift errors while leaving etymological failures largely unaffected.

### 8.1 Qualitative Translation Impact

The study also evaluates downstream translation qualitatively. Without splitting, merged forms are sometimes mistranslated or left untranslated by a general-purpose translator; after splitting, their English meanings become recoverable. Table 9 summarizes the examples reported in the study. This is not a controlled extrinsic benchmark, but it illustrates why Sandhi splitting matters for practical Sinhala NLP (Liyanage and Pushpananda, 2024; Ekanayaka et al., 2023).

Table 9: Qualitative translation examples from the study.

| Input | Without split | After split |
|---|---|---|
| ස්වරාදේශ svarādesha (vowel substitution) | Homeland (wrong) | Vowel substitution |
| විහාරාධිකාර viharādhikāra (temple authorities) | Untranslated | Temple authorities |

## 9 Prototype Implementation

Beyond offline evaluation, the study implements a local prototype for Sinhala Sandhi splitting. The backend is built with FastAPI, selected for its asynchronous request handling and Pydantic-based request validation. The service exposes two endpoints: `/easy` and `/hard`, corresponding to the trained model variants. CORS middleware is configured so that the frontend can communicate with the backend during local deployment.

Table 8: Representative correct and incorrect predictions from the qualitative analysis.

| Input | Prediction | Reference |
|---|---|---|
| විහාරාධිකාර viharādhikāra (temple authorities) | විහාර අධිකාර vihāra adhikāra (temple authority) | විහාර අධිකාර vihāra adhikāra (temple authority) |
| කලායතනය kalāyatanaya (art institute) | කලා ආයතනය kalā āyatanaya (art institution) | කලා ආයතනය kalā āyatanaya (art institution) |
| දකුණත dakunata (right hand) | දකුණ අත dakuna ata (right hand) | දකුණු අත dakunu ata (right hand) |
| කම්මල kammala (smithy) | කම් මල kam mala (work flower) | කම් හල kam hala (work hall/place) |

The frontend is implemented in React. It allows a user to submit a Sinhala Sandhi word and view the predicted split, romanized output, and English translation. The transliteration module uses the same character-level mapping described in the preprocessing pipeline, while the translation component is used to make the output more readable for users who do not read Sinhala. Model weights, tokenizer objects, vocabulary mappings, and sequence length settings are serialized so that the deployed inference pipeline uses the same configuration as training. The thesis implementation also uses NumPy and scikit-learn utilities during data processing and splitting.

The prototype is included as an implementation demonstration rather than as a user study. The study reports functional validation in a local environment, but it does not evaluate usability, latency, or accuracy under real world deployment conditions. For this reason, the prototype supports the practical feasibility of the approach but does not replace the quantitative benchmark.

## 10 Conclusion

This paper frames Sinhala Sandhi splitting as character-level sequence-to-sequence transduction and establishes the first neural benchmark on SandhiLex. The strongest model, a bidirectional LSTM encoder with a unidirectional LSTM decoder, reaches 68.40% exact match and 82.08% character-level accuracy on the hard subset, and 94.00% on the full affixational subset. A key finding is that Sinhala Sandhi is not one problem but several with different tractability profiles: affixational Sandhi is solvable by scaling data alone, while etymological Sandhi remains unsolved by added context or capacity, since the required information is absent from the surface form, pointing toward lexical retrieval, not sequence transduction, as the more promising direction here.

Future work should expand the hard subset, evaluate performance separately by Sandhi type, explore CNN-based and Sandhi-specific Transformer architectures (rather than general-purpose pretrained models) trained directly on SandhiLex, and measure downstream gains in controlled NLP tasks such as machine translation and text-to-speech.

Code, preprocessing scripts, and model implementation for this study are publicly available on GitHub[1]. The SandhiLex dataset itself is not redistributed with this release; it must be obtained separately through a formal academic request to LTRL-UCSC, under the dataset's own academic-use terms (Liyanage and Pushpananda, 2024).

## Limitations

The hard subset contains only about 4,500 pairs, limiting the model's ability to learn rare lexicalized and etymological Sandhi patterns (Liyanage and Pushpananda, 2024). In addition, lexicalized, derivational, and etymological Sandhi are grouped into a single hard category, preventing separate analysis across linguistically distinct Sandhi types.

The encoder-decoder model also lacks attention and relies on greedy decoding during inference. Techniques such as beam search and Sandhi-type conditioned decoding, shown promising in related tasks (Dave et al., 2020), remain future work. Future work could validate predicted splits against a Sinhala lexicon and rerank multiple decoded hypotheses, which may catch some orthographically valid but semantically incorrect splits (see Section 8). Furthermore, evaluation is fully automatic and based on a single reference split, despite the possibility of multiple linguistically valid Sandhi analyses, meaning expert linguistic review would be necessary to assess prediction quality and practical translation impact.

[1] https://github.com/YasasDEK/Sinhala-Sandhi-Splitter

## A Ablation Details

Table 10 reports the full ablation study on the hard subset. Each row changes one component of the full model relative to the bidirectional LSTM baseline (68.40% EM, 82.08% CLA). Encoder directionality is not repeated here, since removing it reduces to the unidirectional LSTM condition already reported in Table 4; see that table and Section 6 for the 26.0-point directionality effect.

Table 10: Ablation study on the hard subset (excluding encoder directionality, reported separately in Table 4). Each row changes one component of the full model.

| Variant | EM | CLA |
|---|---|---|
| Full model: BiLSTM, $d = 512$, Unicode, tuned | **68.40** | 82.08 |
| − native script | 64.00 | **86.27** |
| − latent dimension 512 → 256 | 61.20 | 78.43 |
| − hyperparameter tuning | 64.60 | 78.19 |